\documentclass[letterpaper]{article} 
\usepackage{aaai2026}  
\usepackage{times}  
\usepackage{helvet}  
\usepackage{courier}  
\usepackage[hyphens]{url}  
\usepackage{graphicx} 
\usepackage{natbib}  
\usepackage{caption} 
\usepackage{algorithm}
\usepackage{algorithmic}
\usepackage{latexsym}
\usepackage{amssymb}
\usepackage{amsmath}
\usepackage{amsthm}
\usepackage{booktabs}
\usepackage{enumitem}
\usepackage{graphicx}
\usepackage{multirow}
\usepackage{booktabs}

\usepackage{newfloat}
\usepackage{listings}
\DeclareCaptionStyle{ruled}{labelfont=normalfont,labelsep=colon,strut=off} 
\floatstyle{ruled}
\newfloat{listing}{tb}{lst}{}
\floatname{listing}{Listing}
\title{Modality Bias in LVLMs: \\ Analyzing and Mitigating Object Hallucination via Attention Lens}
\author{
    Haohan Zheng\textsuperscript{\rm 1}\equalcontrib,
    Zhenguo Zhang\textsuperscript{\rm 2}\equalcontrib
}
\affiliations{
    \textsuperscript{\rm 1}Tsinghua Shenzhen International Graduate School, Tsinghua University\\

    \textsuperscript{\rm 2}ShanghaiTech University\\
    
    zhenghh23@mails.tsinghua.edu.cn, zhangzhg2023@shanghaitech.edu.cn \\
    Code is available at \url{https://github.com/zhenghh8/TVAI}.
}

\usepackage{bibentry}

\begin{document}

\maketitle

\begin{abstract}
Large vision-language models (LVLMs) have demonstrated remarkable multimodal comprehension and reasoning capabilities, but they still suffer from severe object hallucination. Previous studies primarily attribute the flaw to linguistic prior caused by the scale mismatch between visual encoders and large language models (LLMs) in LVLMs. Specifically, as current LVLMs are built upon LLMs, they tend to over-rely on textual prompts and internal knowledge of LLMs, generating descriptions inconsistent with visual cues. However, through an in-depth investigation of the hallucinated mechanisms, we empirically reveal a previously overlooked phenomenon: LVLMs may ignore not only visual information but also textual modality during hallucination, a behavior termed as modality bias, which indicates that LVLMs struggle to simultaneously attend to both visual and textual modalities, leading to fragmented understanding of user-provided instructions. Based on this observation, we propose a simple yet effective training-free method to mitigate object hallucination. Concretely, we intervene and adjust the attention weights of textual and visual tokens, balancing cross-modal compatibility for better alignment with user intentions. Furthermore, we adopt a contrastive decoding strategy to reduce the LVLM's overreliance on its parametric knowledge, synergistically enhancing our attention manipulation. Extensive experiments confirm the widespread presence of modality bias in LVLMs. Notably, our method effectively mitigates hallucination across multiple open-source LVLMs and benchmarks, highlighting its generalizability and efficacy.
\end{abstract}


\section{Introduction}\label{sec1}
In recent years, large vision-language models (LVLMs) \cite{bai2023qwen,liu2023visual,zhu2023minigpt,chen2023shikra} have shown unprecedented capabilities and remarkable versatility in multimodal reasoning and human-assistant interaction, attracting widespread attention from the community. Such models are capable of handling interleaved text-image input and have great potential for development in a wide range of domains, such as autonomous driving \cite{cui2024survey,tian2024drivevlm} and healthcare \cite{li2024llava,xu2024multimodal}. Although existing LVLMs have achieved impressive performance, they still suffer from serious object hallucination \cite{li2023evaluating,rohrbach2018object,wu2024logical}. Specifically, LVLMs often exhibit large uncertainty about the presence and properties of objects in user-provided images, generating incorrect hallucinated statements, which severely hampers their deployment in real-world scenarios.

\begin{figure*}[t]
  \centering
    \includegraphics[width=\textwidth]{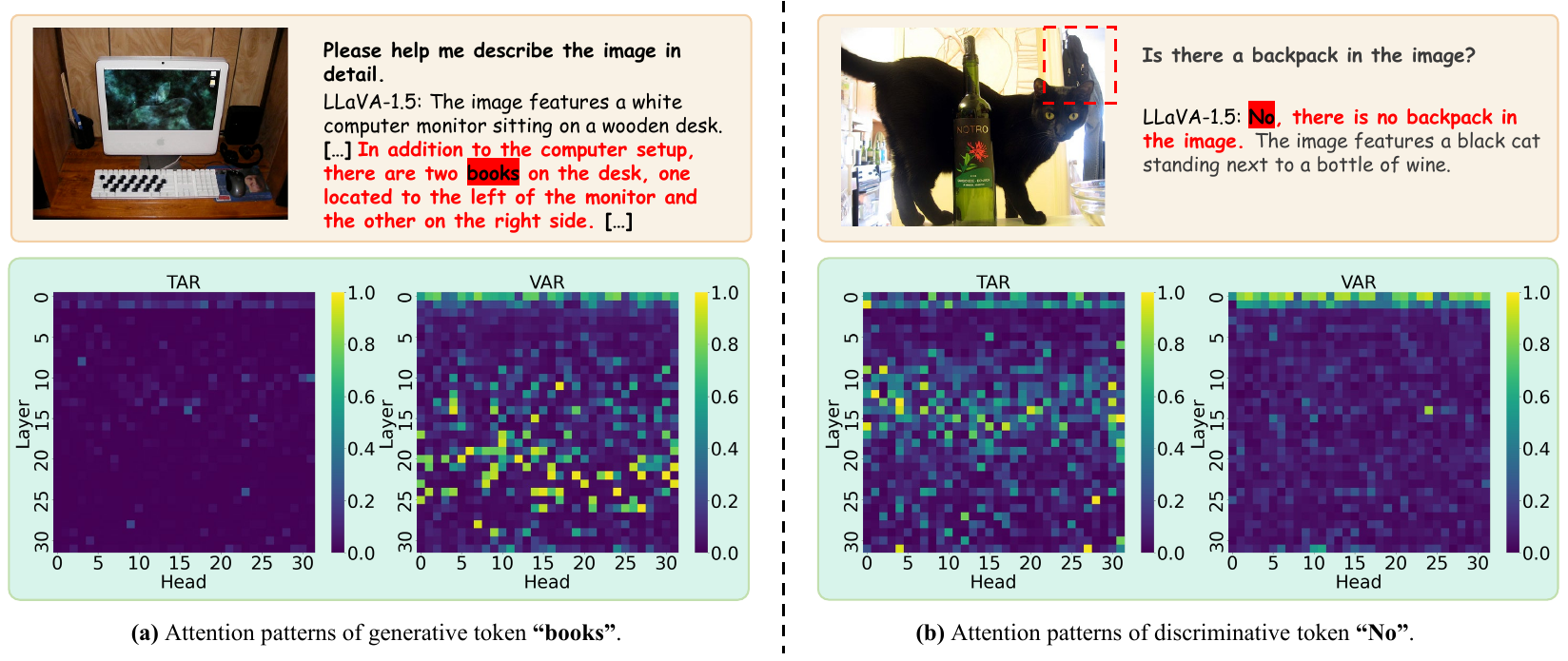}
    \caption{Modality bias in LVLMs. Attention patterns of hallucinated tokens for textual and visual tokens from user-provided instructions, respectively. (a) Generative hallucination. (b) Discriminative hallucination. The hallucinated description is highlighted in red. VAR: visual attention ratio, TAR: textual attention ratio, are defined in Eqs. \ref{eq5}-\ref{eq6}.}
    \label{teaser}
\end{figure*}

Previous works have explored in depth the underlying mechanisms for the occurrence of object hallucination in LVLMs, mostly attributed to fine-grained underrepresentation \cite{chen2024halc,chen2024ict,cho2022fine,an2024agla} and language prior \cite{liu2024paying,chen2024image,jiang2024devils}. Classic network architecture of LVLMs mainly consists of visual encoders and large language models (LLMs) for processing image and text respectively. Existing cross-modal alignment techniques often fail to effectively capture fine-grained semantic associations, leading to limitations in the ability of LVLMs to accurately represent and reason about object properties. Furthermore, the parameter size of LLMs significantly exceeds that of visual encoders, e.g., Vicuna \cite{chiang2023vicuna} contains over ten times the number of parameters compared to CLIP \cite{radford2021learning}. This discrepancy leads to an over-reliance on linguistic knowledge rather than visual inputs, ultimately resulting in generated information that is misaligned with the corresponding image. However, these studies have primarily focused on the insufficiency of visual information as a cause of object illusions, overlooking the critical role that textual inputs play in determining the quality of model-generated outputs.

In addition, significant efforts have been devoted to developing strategies to mitigate hallucination in LVLMs, primarily through interventions across three distinct phases: the training, inference, and post-processing stages. The most straightforward approaches involve intervention during the post-processing stage \cite{zhou2023analyzing,chuang2024lookback}, where post-hoc corrections are applied directly to the hallucinated outputs. During the training phase, these methods \cite{li2024vlfeedback,gunjal2024detecting} incorporate additional curated, high-quality annotated data to re-training or fine-tune the LVLMs, enhancing the instruction-following capabilities. The currently prevalent paradigm focuses on inference-stage interventions \cite{liu2024reducing,zhao2024looking,chen2024halc,chuang2023dola}, employing diverse decoding strategies to effectively mitigate hallucination by penalizing language bias. However, these methods entail substantial incremental costs, including the need for additional data annotation, increased inference time, and the computational overhead associated with re-training or integrating external tools and models.

In this paper, we delve deep into how object hallucination can be effectively mitigated in LVLMs without introducing significant additional costs. Our study is grounded in an empirical observation that LVLMs exhibit two distinct attention patterns when generating object-related hallucinated descriptions. The first pattern involves the LVLMs mainly focusing on visual representation from the instructions during token generation, while the second pattern reveals a bias toward textual information. In contrast to previous studies that emphasize language prior \cite{wang2024mllm,leng2024mitigating,liu2024survey}, we demonstrate that LVLMs can also display insufficient attention to textual information, which contributes to hallucination. It is noteworthy that an intriguing phenomenon emerges, which we refer to as \emph{modality bias} in LVLMs, as illustrated in Figure \ref{teaser}. When generating hallucinated objects, the model focuses primarily on visual inputs and disregards textual information. Conversely, when generating incorrect responses regarding the presence or absence of objects, the model predominantly relies on textual representations while neglecting visual cues. This finding aligns with intuitive expectations. When a user inquires about the existence of an object, the model should prioritize the specific details of the object’s description provided in the textual input. Similarly, when generating descriptions of an image, the model should exhibit a stronger bias toward the visual information supplied by the user.

We refer to the object hallucination resulting from these two distinct attention patterns as generative and discriminative hallucinations, respectively. Specifically, generative hallucination occurs when LVLMs generates an object description that is inconsistent with the image, while discriminative hallucination arises when the model, in response to a query about the existence of a specific object, generates a binary answer, i.e. yes or no, that contradicts the actual scenario. For generative hallucination, it demonstrates a greater emphasis on visual representation, whereas for discriminative hallucination, the model exhibits a stronger reliance on textual information. However, LVLMs can only generate accurate, instruction-following responses and avoid hallucination if they effectively attend to both the textual and visual representations.

Inspired by the above analysis, an intuitive strategy to mitigate object hallucination is to ensure that LVLMs align the densities of textual and visual information during token generation, thereby eliminating unimodal bias. Therefore, we propose TVAI, a training-free method designed to reduce object hallucination via \underline{T}extual and \underline{V}isual \underline{A}ttention \underline{I}ntervention. We focus on the self-attention mechanisms within the decoding layers of LVLMs, manipulating the attention weights assigned to instruction tokens during the inference stage. Concretely, to alleviate generative hallucination, we augment the attention weights of text tokens along their original direction prior to token generation. Similarly, we recalibrate the attention weights of image tokens, enabling the LVLMs to significantly reduce discriminative hallucination. By intervening in the attention matrix, the model achieves a more balanced hidden state, thereby effectively reducing the object hallucination. From a broader perspective, the proposed method represents a step toward the development of a general-purpose assistant for future applications. TVAI enhances the ability of LVLMs to integrate fine-grained visual and textual information in a more comprehensive manner, moving beyond superficial or disjointed representations. In contrast to prior studies, our approach does not incur significant additional computational costs. Moreover, we reinterpret object hallucination through a novel perspective, moving beyond attributing them solely to language bias. Since TVAI intervenes exclusively with attention weights during the inference phase, it is universally applicable to any decoding strategy across all LVLMs.

We conducted extensive experiments across four popular LVLMs, with the results demonstrating the superior capability of our approach in mitigating hallucination. Our experiments leverage CHAIR and POPE benchmarks to evaluate LVLMs' accuracy in long-sequence generation and visual question answering (VQA) tasks, respectively, and MMBench for general capability, comprehensively assessing TVAI’s anti-hallucination efficacy. Overall, our main contributions are summarized as follow:
\begin{enumerate}
    \item We identify two distinct attention patterns during hallucination in LVLMs, image-centered or text-centered, revealing that modality bias is a primary driver of object hallucination.
    \item A training-free approach to reduce hallucination is proposed, which manipulates attention weights during the inference phase to ensure the model remains more aligned with the user’s instructions.
    \item Substantial experiments demonstrate that TVAI effectively mitigates hallucination without incurring additional data requirements and significant computational overhead.
\end{enumerate}

\section{Related Work}\label{sec2}
Hallucination refer to the inconsistency between the generated content and the user-provided instruction. Given that hallucination significantly impede the grounded application of LVLMs in real-world scenarios, extensive researches have been conducted to investigate the main causes of this phenomenon \cite{lee2021deduplicating,huang2025survey,peng2023check}. The most immediate reason is the distributional bias between textual and visual modalities, coupled with the model’s insufficient understanding of the relationships across these modalities. Consequently, richer and meticulously curated datasets have been integrated into LVLMs \cite{gunasekar2023textbooks} for re-training or fine-tuning, enhancing semantic consistency across modalities. While these methods demonstrate efficacy in reducing hallucination, they are time-intensive, computationally demanding, and incur substantial additional costs for data annotation and training. In addition to direct enhancements to LVLMs, hallucination can also be mitigated through post-processing methods \cite{zhou2023analyzing,chuang2024lookback}. These approaches typically employ external tools to detect hallucination, followed by the integration of a revisor to refine the generated outputs, also substantially increasing inference costs.

Recently, training-free strategies have gained spark interest among researchers due to their ability to effectively mitigate hallucination without requiring additional alignment training or external tools. OPERA \cite{huang2024opera} found an unusual knowledge aggregation pattern in the inference processes of LVLMs, a phenomenon strongly correlated with the occurrence of hallucination. Based on this observation, they proposed a technique to monitor and mitigate hallucination, which is specifically tailored to the beam search decoding strategy. VCD \cite{leng2024mitigating} demonstrated that existing LVLMs are excessively influenced by statistical bias and language priors, which are main contributors to hallucination. To address this, they introduced a visual contrastive decoding method based on the nucleus sampling strategy, aimed at mitigating language priors and ultimately reducing hallucination. A concept analogous to our work is presented in PAI \cite{liu2024paying}, which observed that despite image tokens constituting a substantial proportion, e.g. 576 image tokens in LLaVA-1.5 \cite{liu2024improved}, they received low attention during the inference process. In light of this, PAI proposed a method to directly augment the attention weights of image tokens during the decoding process, achieving notable performance in hallucination reduction. 

Unlike prior works, we identify the presence of two distinct attention patterns in LVLMs during hallucination events, as illustrated in Figure \ref{teaser}. These patterns lead to hallucination when LVLMs disproportionately focus on either textual or visual representations. Inspired by the preceding analysis, we introduce TVAI, a training-free attention manipulation strategy. TVAI demonstrates exceptional hallucination mitigation performance by eliminating unimodal bias, thereby ensuring a more balanced attention distribution across textual and visual information.

\begin{figure*}[t]
    \centering
    \includegraphics[width=0.9\textwidth]{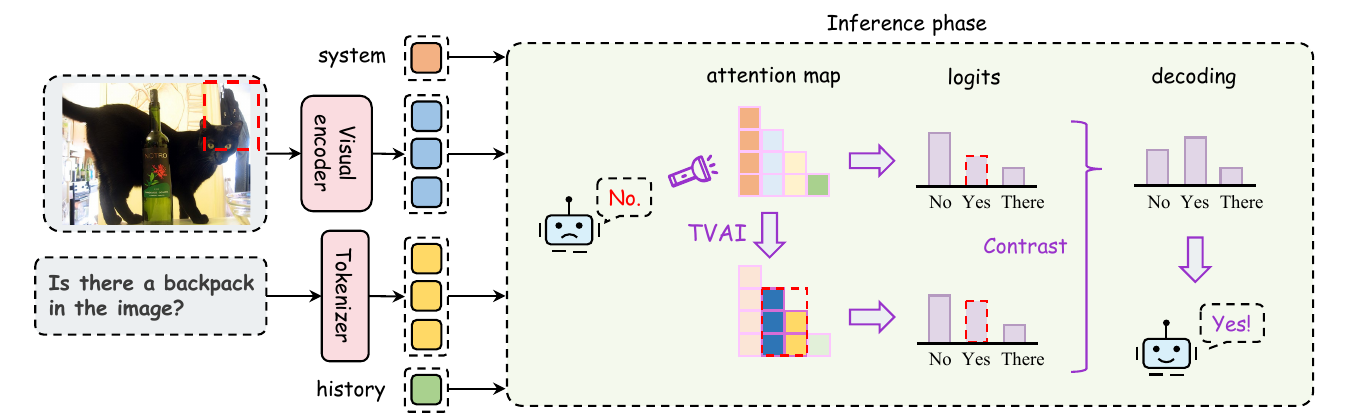}
    \caption{Overview of our TVAI (Textual and Visual Attention Intervention). The TVAI framework adjusts visual-textual token attention weights to balance cross-modal compatibility and align with user intentions, while integrating contrastive decoding to reduce LVLMs’ overreliance on parametric knowledge and synergistically enhance attention manipulation.}
    \label{tvai_framework}
\end{figure*}

\section{Methodology}\label{sec3}

\subsection{Preliminaries}\label{sec3.1}
LVLMs typically comprise three core components: a visual encoder, a projector, and a language decoder. During the reasoning process, the visual encoder encodes user-provided images into image tokens, which are subsequently mapped to the textual representation space via the projector. These image tokens are then concatenated with text tokens and fed into the language decoder to generate the corresponding responses. Current state-of-the-art language decoders are predominantly composed of $n_L$ transformer blocks \cite{vaswani2017attention}, with the multi-head self-attention mechanism playing a critical role in generating responses that accurately adhere to user instructions. We define each block to contain $n_H$ heads, and the self-attention operation of the $h$-th head in the $l$-th layer can be formulated as follows:
\begin{equation}
    \mathbf{A}_{l,h}=\frac{\mathbf{Q}_{l,h}\mathbf{K}_{l,h}^{\top}}{\sqrt{d_k}}
    \label{eq1}
\end{equation}
\begin{equation}    
\mathbf{O}_{l,h}=\text{softmax}(\mathbf{A}_{l,h})\mathbf{V}_{l,h}
    \label{eq2}
\end{equation}

In the $l$-th layer transformer block, each attention head $h$ is associated with its own queries $\mathbf{Q}_{l,h}$, keys $\mathbf{K}_{l,h}$, and values $\mathbf{V}_{l,h}$, $\mathbf{Q}_{l,h},\mathbf{K}_{l,h},\mathbf{V}_{l,h}\in\mathbb{R}^{n \times d_k}$, where $n$ represents the length of the input sequence and $d_k$ denotes the dimension of the hidden state. The attention matrix $\mathbf{A}_{l,h}\in \mathbb{R}^{n \times n}$ is computed as the product of queries $\mathbf{Q}_{l,h}$ and keys $\mathbf{K}_{l,h}$, where each row represents the attention weights of a token relative to all other tokens during information integration. Each row of $\mathbf{V}_{l,h}$ encapsulates the representation embedded in a token, and the product of the attention matrix $\mathbf{A}_{l,h}$ and $\mathbf{V}_{l,h}$ computes the weighted sum $\mathbf{O}_{l,h}$ of all tokens for each individual token. In other words, during the inference phase of LVLMs, each generated token is a feature fusion derived from the attention weights applied to the user-provided text tokens $\mathbf{X}^{T}$, image tokens $\mathbf{X}^{V}$, and historically generated tokens $\mathbf{X}^{H}$. In a multi-head attention mechanism, the outputs of all attention heads are concatenated and linearly projected using a weight matrix $\mathbf{W}^{O}$ to produce the final result. After multiple iterations of the self-attention operation, the final hidden state $\mathbf{H}$ is derived. During the generation of the $k$-th token, the final hidden state $\mathbf{H}_k$ is mapped to the vocabulary space via a fully connected layer $\mathbf{W}^H$, yielding the conditional probability distribution $p\in \mathbb{R}^v$, which can be expressed as follows:
\begin{equation}
    \mathbf{H}_{1\sim k}=\text{concat}(\mathbf{O}_{n_L,1},\mathbf{O}_{n_L,2},\dots,\mathbf{O}_{n_L,n_H})\mathbf{W}^O
    \label{eq3}
\end{equation}
\begin{equation}
    p(y_k|y_{<k})=\text{softmax}(\mathbf{H}_k\mathbf{W}^H)
    \label{eq4}
\end{equation}
where $v$ denotes the size of the vocabulary and $y_k$ represents the token generated at the $k$-th step.

\subsection{Generative and Discriminative Hallucinations}\label{sec3.2}
When generating the $k$-th token, the input text $\mathbf{X}^T=[x_{t_1},x_{t_2}, \dots, x_{t_T}]$, image $\mathbf{X}^V=[x_{v_1},x_{v_2}, \dots, x_{v_V}]$, and historically generated information $\mathbf{X}^H$ are fed into the language decoder. To empirically investigate the two distinct attention patterns associated with generative and discriminative hallucinations, as described in Section \ref{sec1}, we introduce two metrics: the textual attention ratio (TAR) and the visual attention ratio (VAR). These metrics quantify the attention allocated to text and image tokens from instruction, respectively, during token generation, and are defined by the following equations:
\begin{equation}    
\mathbf{TAR}_{l, h}=\sum_{i=t_1}^{t_{n_T}}\mathbf{A}_{l, h}(k,i)
    \label{eq5}
\end{equation}
\begin{equation}    
\mathbf{VAR}_{l, h}=\sum_{j=v_1}^{v_{n_V}}\mathbf{A}_{l, h}(k,j)
    \label{eq6}
\end{equation}

TAR and VAR quantify the cumulative attention assigned by the newly generated $k$-th token to the input text and image tokens, respectively. In Figure \ref{teaser}, we provide two illustrative examples to demonstrate the distinct attention patterns: vision-centered and text-centered. To more comprehensively validate the prevalence of this phenomenon in LVLMs, we conduct experiments on the COCO dataset \cite{lin2014microsoft} using two widely recognized open-source models, i.e. LLaVA-1.5 and Qwen-VL-Chat. Detailed experimental configurations are provided in the Supplementary Material. We compute the TAR and VAR for hallucinated tokens during both generative and discriminative hallucination events, respectively. The calculations are then averaged to derive the final metrics, as illustrated in Figure \ref{overall_pattern}. The experimental results demonstrate the presence of two distinct attention patterns in LVLMs: one predominantly focus on visual representation and the other primarily on textual information, reflecting a tendency of the model to process modalities independently rather than in an integrated manner. This phenomenon aligns with the intuitive expectation that LVLMs should allocate balanced and modality-aware attention to both textual and visual instructions, as neglecting either modality may contribute to the occurrence of semantic inconsistency and hallucination.

\subsection{Textual and Visual Attention Intervention}\label{sec3.3}
Motivated by the empirical findings from the above analysis, a straightforward idea is to adjust the TAR and VAR during the inference phase, thereby mitigating the model’s tendency to disproportionately favor a single modality during token generation. Therefore, our goal is to identify reliable directions and optimal excitation locations for manipulating attention weights. \cite{jiang2024devils} enhanced visual attention scores by augmenting with the average absolute value of all attention heads within the same layer, improving the alignment of different attention heads toward the same image region. \cite{chen2024ict} applied Gaussian blur to image regions to derive an activation shift vector, then an additional classifier is trained to determine which attention heads require activation intervention. In contrast, we posit that the visual encoder and language decoder of LVLMs have been pretrained on extensive datasets, endowing them with robust comprehension capabilities for image and text inputs, respectively. Consequently, by directly augmenting the attention weights of image and text tokens along their original attention directions, the model can be guided toward more reliable responses. This approach mitigates unimodal bias through manipulating textual and visual attention directly, can be expressed as follows:
\begin{equation}
    \mathbf{A}^{\prime}_{l, h}(k,i)=\mathbf{A}_{l, h}(k,i) + \alpha |\mathbf{A}_{l, h}(k,i)|, i=t_1 \sim t_{n_T}
    \label{eq7}
\end{equation}
\begin{equation}
    \mathbf{A}^\prime_{l, h}(k,j)=\mathbf{A}_{l, h}(k,j) + \beta |\mathbf{A}_{l, h}(k,j)|, j=v_1 \sim v_{n_V}
    \label{eq8}
\end{equation}
where $\alpha$ and $\beta$ govern the step size of the TAR and VAR excitation, respectively. It is important to note that we intervene in attention weights before softmax operation. 

Furthermore, when the attention distribution is non-redundant, augmenting the attention weights of both text and image tokens fails to achieve the intended effect, as the majority of the attention is primarily concentrated on the user instructions. Conversely, in the presence of redundant attention, an attention sink pattern becomes evident in LVLMs \cite{liu2024paying,xiao2023efficient,darcet2023vision}. In shallow layers, this phenomenon is generally absent, as the model prioritizes encoding semantically rich information. In deeper layers, however, where the hidden states progressively stabilize, the attention sink pattern emerges concurrently, signifying the occurrence of attentional redundancy. Therefore, our attention manipulation is applied specifically after the emergence of the attention sink phenomenon, effectively enhancing the influence of textual and visual tokens during generation. Using Qwen-VL-Chat as an example, Figure \ref{qwen_pattern} displays the average values of the TAR and VAR across each decoder layer. The optimal excitation location of the textual and visual attention intervention is determined by the divergence of attention patterns between generative and discriminative hallucinations.

\begin{table}[h]
\centering
\setlength{\tabcolsep}{3pt} 
\begin{tabular}{ccccccc}
    \toprule
    \multirow{2}{*}{Method} & \multicolumn{2}{c}{LLaVA-1.5} & \multicolumn{2}{c}{MiniGPT-4} & \multicolumn{2}{c}{Shikra}\\
    \cmidrule(lr){2-3} \cmidrule(lr){4-5} \cmidrule(lr){6-7}
     & $C_S\downarrow$ & $C_I\downarrow$ & $C_S\downarrow$ & $C_I\downarrow$ & $C_S\downarrow$ & $C_I\downarrow$ \\
    \midrule 
    Vanilla & 47.6 & 13.7 & 29.6 & 10.0 & 46.0 & 14.2\\
    \midrule
    OPERA & 45.1 & 13.1 & 25.4 & 9.6 & 39.6 & 12.5\\
    VCD & 59.1 & 17.9 & 41.2 & 13.8 & 56.4 & 15.1\\ 
    PAI & 24.3 & 7.1 & 23.5 & 8.6 & 38.8 & 10.1\\
    \midrule
    TAI w/o Contrast & 35.8 & 9.1 & 25.0 & 9.7 & 40.8 & 11.8\\
    VAI w/o Contrast & 36.8 & 10.3 & 22.4 & 7.8 & 44.8 & 12.8\\
    TVAI w/o Contrast & 27.6 & 6.8 & 18.2 & 6.6 & \textbf{34.4} & \textbf{8.7}\\
    TAI & 33.4 & 8.9 & 23.2 & 8.9 & 39.2 & 11.2\\
    VAI & 35.6 & 10.0 & 21.8 & 8.3 & 39.6 & 12.1\\
    TVAI & \textbf{22.4} & \textbf{6.3} & \textbf{15.4} & \textbf{5.2} & 35.0 & 9.4\\
    \bottomrule
\end{tabular}
\caption{CHAIR hallucination evaluation results on three open-source LVLMs. TAI and VAI are simplified variants of TVAI (only textual/visual attention intervened). Values are in percentage (\%) with 1 decimal place; best results (lowest $C_S, C_I$) are bolded.}
\label{tab1}
\end{table}

\subsection{Contrast Decoding}\label{sec3.4}
Fig. \ref{tvai_framework} presents an overview of our proposed TVAI framework. As shown, the conditional probability of the correct token increases when textual and visual attention intervention (TVAI) is applied, as LVLMs are more effectively guided by user instructions during the generation process. However, due to pretraining on massive-scale data, LVLMs still exhibit overreliance on their internal parametric knowledge, which may lead to hallucination. To address this, we propose a contrastive decoding strategy that reinforces the effect of TVAI during inference, reduces the model's dependence on incorrect commonsense priors, and further enhances its hallucination mitigation capability. The conditional probability of the final LVLMs at each generation step are shown in the following equation: 
\begin{equation}
    p_{final}=\gamma \cdot p^\prime(y_k|y_{<k}) + (1 - \gamma) \cdot p(y_k|y_{<k})
    \label{eq9}
\end{equation}
where $\gamma (\gamma > 1)$ controls the strength of the contrastive decoding, $p^\prime(y_k \mid y_{<k})$ denotes the token probability generated from the modified attention map $\mathbf{A}^\prime_{l,h}$ after adopting TVAI, and $p(y_k \mid y_{<k})$ corresponds to the original output derived from the vanilla attention weight $\mathbf{A}_{l,h}$.

\section{Experiments}\label{sec4}

\subsection{Experimental Setup}\label{sec4.1}
\textbf{Models.} To validate the effectiveness and generalization capability of our method, we employ four popular open-source LVLMs, i.e., LLaVA-1.5 \cite{liu2024improved}, Qwen-VL-Chat \cite{bai2023qwen}, MiniGPT-4 \cite{zhu2023minigpt}, and Shikra \cite{chen2023shikra}. 

\noindent\textbf{Baseline.} As baseline models, we employ the vanilla LVLMs and three state-of-the-art hallucination mitigation methods: OPERA \cite{huang2024opera}, VCD \cite{leng2024mitigating}, and PAI \cite{liu2024paying}. Details of the implementation can be found in the Supplementary Material. 

\noindent\textbf{Benchmark and Metrics.} We selected CHAIR \cite{rohrbach2018object} and POPE \cite{li2023evaluating}, two established datasets specifically tailored to hallucination evaluation, as our benchmarks to rigorously and fairly validate the effectiveness of our method. CHAIR is employed to evaluate object hallucination in the image captioning task. This metric quantifies the proportion of objects mentioned in a generated caption that are not present in the ground-truth annotations. CHAIR comprises two sub-metrics as described in Eqs. \ref{eq10}-\ref{eq11}, designed to assess hallucination at the instance-level and sentence-level, respectively. Additionally, we utilize MMBench \cite{liu2024mmbench} to evaluate the general capability of our model, ensuring a comprehensive validation of both hallucination mitigation efficacy and overall performance.

\begin{equation}
    C_S=\frac{|\{\text{hallucinated captions}\}|}{|\{\text{all captions}\}|}
    \label{eq10}
\end{equation}
\begin{equation}
    C_I=\frac{|\{\text{hallucinated objects}\}|}{|\{\text{all generated objects}\}|}
    \label{eq11}
\end{equation}

A smaller CHAIR metric indicates better hallucination mitigation, assessing caption-based object hallucination. In contrast, POPE evaluates LVLMs’ discriminative object recognition, with accuracy and F1 (following \cite{liu2024paying}). For MMBench, we use overall average accuracy, macro average precision, recall, and F1 to measure its Chinese and English subsets respectively.

\subsection{Experimental Results}

\noindent\textbf{Long-sequence generation task.} This task requires LVLMs to produce comprehensive and accurate descriptions of input images. We evaluate our proposed TVAI on three widely adopted and open-source LVLMs: LLaVA-1.5, MiniGPT-4, and Shikra. The results, summarized in Table \ref{tab1}, demonstrate that TVAI consistently delivers superior performance across all models. In particular, LVLMs augmented with TVAI exhibit a substantial reduction in both instance-level and sentence-level object hallucinations, as measured by CHAIR metrics. Finally, Table \ref{tab1} includes multiple configurations of TVAI under varying settings. Across all variants, TVAI consistently achieves state-of-the-art performance in hallucination mitigation while preserving, and in some cases slightly improving, the general ability compared to the original models. These results underscore the effectiveness and robustness of the proposed intervention framework in enhancing the fidelity of LVLM-generated descriptions.

\begin{table}[h]
\centering
\setlength{\tabcolsep}{3pt} 
\begin{tabular}{ccccccc}
    \toprule
    \multirow{2}{*}{Method} & \multicolumn{2}{c}{LLaVA} & \multicolumn{2}{c}{MiniGPT-4} & \multicolumn{2}{c}{Shikra}\\
    \cmidrule(lr){2-3} \cmidrule(lr){4-5} \cmidrule(lr){6-7}
     & Acc. & F1 & Acc. & F1 & Acc. & F1 \\ 
    \midrule 
    Vanilla & 79.3 & 81.1 & 70.6 & 71.1 & 76.2 & 78.4\\
    \midrule 
    OPERA & 79.8 & 80.9 & 70.7 & 70.8 & 77.2 & 77.9\\
    VCD & 78.7 & 81.3 & 71.3 & 71.2 & 77.5 & 78.5\\
    PAI & 80.3 & 81.7 & 71.0 & 70.9 & 78.1 & 78.6\\
    TVAI & \textbf{81.2} & \textbf{81.8} & \textbf{71.7} & \textbf{71.5} & \textbf{78.7} & \textbf{79.8}\\
    \bottomrule
\end{tabular}
\caption{POPE hallucination evaluation results in adversarial configuration on three open-source LVLMs (LLaVA-1.5, MiniGPT-4, Shikra). Values are in percentage (\%) with 1 decimal place; best results are bolded.}
\label{tab2}
\end{table}

\noindent\textbf{Visual question answering task.} This task requires LVLMs to demonstrate strong comprehension of user-provided images and the ability to handle questions that are closely tied to visual content. Following the same evaluation framework used in the image captioning task, we assess both previous state-of-the-art methods and our proposed TVAI across three widely used LVLMs: LLaVA-1.5, MiniGPT-4, and Shikra. The results are presented in Table \ref{tab2}. As the table shows, our TVAI method consistently achieves superior performance on the visual question answering (VQA) task. Specifically, we use the POPE benchmark to quantify the effectiveness of each model. POPE evaluates performance using two metrics: accuracy and F1 score, which together reflect a model’s ability to correctly detect the presence or absence of specified objects in images. POPE provides three benchmark settings—random, popular, and adversarial. Among the three, the adversarial setting poses the greatest challenge because it includes absent objects that are commonly associated with the visual context. This exploits LVLMs' tendency to rely on parametric knowledge or learned priors rather than grounding their answers in the actual image. We report only the results under the adversarial setting in Table \ref{tab2}. The results clearly demonstrate that models equipped with TVAI achieve significant improvements over vanilla LVLMs and other advanced hallucination mitigation strategies.

\begin{figure}[h]
    \centering
    \includegraphics[width=0.4\textwidth]{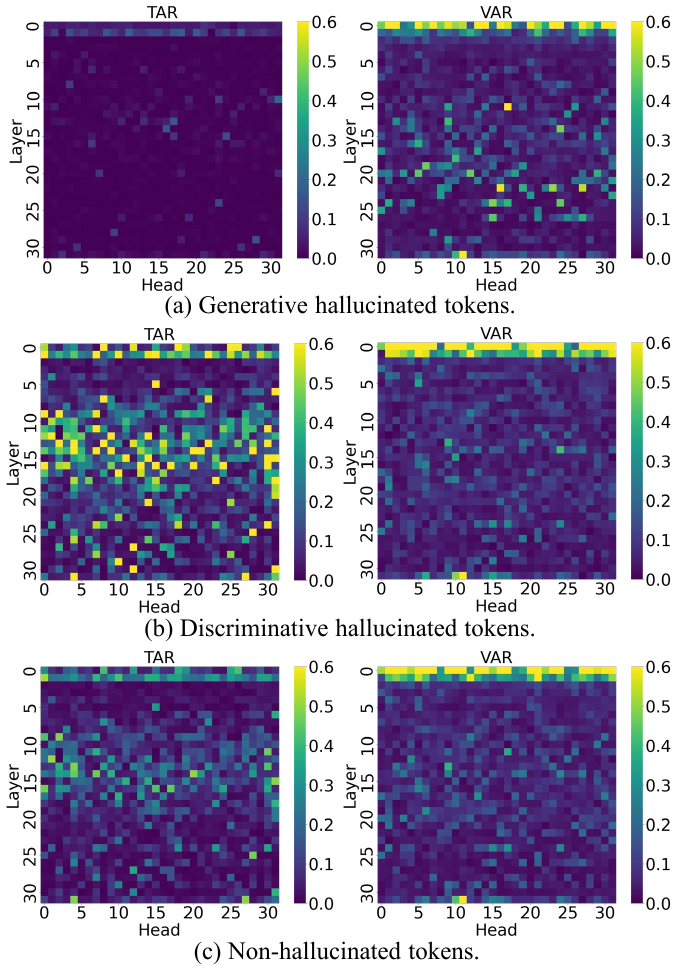}
    \caption{TAR and VAR attention patterns across three token types of LLaVA-1.5, highlighting closer alignment in non-hallucinated tokens without modality bias.}
    \label{overall_pattern}
\end{figure}

\begin{table}[h]
\centering
\begin{tabular}{cccccc}
\toprule
\multicolumn{2}{c}{Setting} & Accuracy & Precision & Recall & F1 \\
\midrule
\multirow{2}{*}{EN} 
    & Vanilla & \textbf{73.3} & \textbf{75.8} & \textbf{73.1} & \textbf{73.6} \\
    & TVAI & 73.1 & 74.5 & 73.1 & 73.3 \\
\midrule
\multirow{2}{*}{CN} 
    & Vanilla & 67.7 & 70.9 & 67.5 & 68.0 \\
    & TVAI & \textbf{67.8} & \textbf{71.0} & \textbf{67.8} & \textbf{68.1} \\
\bottomrule
\end{tabular}
\caption{Performance results on MMBench (EN: English subset; CN: Chinese subset). Values are in percentage (\%) with 1 decimal place; best results are bolded.}
\label{mmbench}
\end{table}

\begin{table*}[t]
\renewcommand{\arraystretch}{1.1}  
\setlength{\tabcolsep}{2pt}       
\centering
    \begin{minipage}{0.32\textwidth}
    \centering
    \begin{tabular}{ccc|ccc}
    \toprule
    $\alpha$ & $\beta$ & $\gamma$ & CHAIR$_\text{S}$ & CHAIR$_\text{I}$ & F1 \\
    \midrule
    - & 0.5 & - & 0.368 & 0.103 & 0.774 \\
    0.7 & 0.5 & - & 0.460 & 0.119 & 0.758 \\
    0.8 & 0.5 & - & 0.396 & 0.106 & 0.774\\ 
    0.9 & 0.5 & - & 0.348 & 0.086 & 0.759\\ 
    \textbf{0.93} & \textbf{0.5} & \textbf{-} & \textbf{0.276} & \textbf{0.068} & \textbf{0.734}\\ 
    0.96 & 0.5 & - & 0.166 & 0.060 & 0.688\\ 
    1.0 & 0.5 & - & 0.078 & 0.046 & 0.633\\ 
    \bottomrule
    \end{tabular}
    \end{minipage}
    \hfill
    \begin{minipage}{0.32\textwidth}
    \centering
    \begin{tabular}{ccc|ccc}
    \toprule
    $\alpha$ & $\beta$ & $\gamma$ & CHAIR$_\text{S}$ & CHAIR$_\text{I}$ & F1 \\
    \midrule
    0.93 & 0.5 & 0.5 & 0.394 & 0.099 & 0.780\\
    0.93 & 0.5 & 0.7 & 0.354 & 0.084 & 0.772\\
    0.93 & 0.5 & 0.9 & 0.296 & 0.089 & 0.744\\
    0.93 & 0.5 & 1.0 & 0.276 & 0.068 & 0.734\\
    \textbf{0.93} & \textbf{0.5} & \textbf{1.1} & \textbf{0.224} & \textbf{0.063} & \textbf{0.711}\\
    0.93 & 0.5 & 1.3 & 0.192 & 0.079 & 0.683\\
    0.93 & 0.5 & 1.5 & 0.166 & 0.071 & 0.652\\
    \bottomrule
    \end{tabular}
    \end{minipage}
    \hfill
    \begin{minipage}{0.32\textwidth}
    \centering
    \begin{tabular}{ccc|ccc}
    \toprule
    $\alpha$ & $\beta$ & $\gamma$ & CHAIR$_\text{S}$ & CHAIR$_\text{I}$ & F1 \\
    \midrule
    0.8 & 0.4 & 0.5 & 0.302 & 0.090 & 0.692 \\
    0.8 & 0.4 & 0.7 & 0.274 & 0.080 & 0.686 \\
    0.8 & 0.4 & 0.9 & 0.218 & 0.072 & 0.682 \\
    0.8 & 0.4 & 1.0 & 0.182 & 0.066 & 0.669 \\
    \textbf{0.8} & \textbf{0.4} & \textbf{1.1} & \textbf{0.154} & \textbf{0.052} & \textbf{0.652} \\
    0.8 & 0.4 & 1.3 & 0.112 & 0.057 & 0.613 \\
    0.8 & 0.4 & 1.5 & 0.080 & 0.057 & 0.604 \\
    \bottomrule
    \end{tabular}
    \end{minipage}
\caption{Ablation study on the manipulation strength of textual and visual attention ($\alpha$, $\beta$) and the influence of contrast decoding intensity ($\gamma$). Results are reported for LLaVA-1.5 (left and middle) and MiniGPT-4 (right). The configuration of the main experimental results in this study is highlighted as \textbf{bold font}.}
\label{tab3}
\end{table*}

\noindent\textbf{General capability evaluation.}
Beyond measuring TVAI's hallucination mitigation capability across the aforementioned tasks, we also evaluated whether TVAI impacts the LVLMs' general capability using the MMBench benchmark. As shown in Table \ref{mmbench}, TVAI effectively preserves the inherent capabilities of the vanilla model and even yields improvements in some cases. This can be attributed to TVAI enabling the model to simultaneously attend to information from different modalities in user instructions, thereby enhancing the model's perceptual ability. These results indicate that modality bias in LVLMs significantly impairs perceptual capability.

\subsection{Ablation Studies}\label{sec4.3}
A consistent performance gap persists between simplified variants and full TVAI (Table~\ref{tab1}), which jointly manipulates textual and visual attention—highlighting multimodal intervention’s complementarity and synergy. Since both visual content and textual instructions ground LVLM outputs, single-modality enhancement may amplify modality bias. To quantify this, controlled experiments varied textual/visual attention intervention strengths (LLaVA-1.5, long-sequence generation; Table~\ref{tab3} left). Fixing visual strength $\beta$ and increasing textual $\alpha$ improved hallucination reduction, but $\alpha=1$ caused minimal hallucinations yet F1 mismatch (overcorrection risk). Thus, main experiments use $\alpha=0.93$, $\beta=0.5$ for LLaVA-1.5. Optimal hyperparameters for other models (provided in Supplementary Material) vary by LVLMs due to architectural/training/data differences (e.g., 576 vs. 32 visual tokens in LLaVA-1.5 vs. MiniGPT-4).

\begin{figure}[h]
    \centering
    \includegraphics[width=0.45\textwidth]{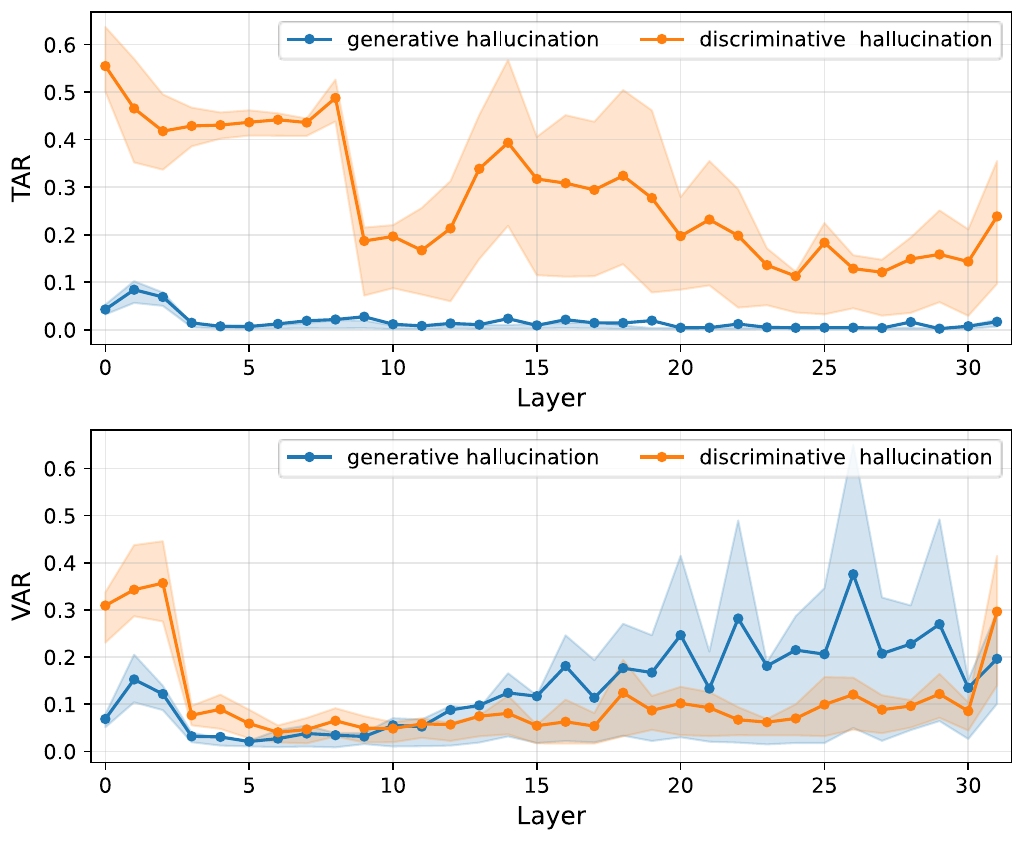}
    \caption{The average textual (above) and visual (bottom) attention ratios of hallucinated tokens in each layer of Qwen-VL-Chat. The shaded regions represent the upper and lower quartiles of the attention values.}
    \label{qwen_pattern}
\end{figure}

Additionally, we investigate TVAI’s contrast decoding module, which amplifies attention manipulation during inference. Using LLaVA-1.5 and MiniGPT-4 on image captioning, we evaluated varying contrast decoding parameter $\gamma$ (Table~\ref{tab3}, middle/right). Results confirm its advantage: hallucination mitigation improves with increasing $\gamma$. However, as TVAI operates purely at inference (no base LVLMs re-training/fine-tuning), excessively high $\gamma$ can deviate from the model’s original trajectory, disrupting precision-recall balance and reducing output reliability. Thus, selecting an appropriate $\gamma$ is critical for maintaining TVAI’s stability and effectiveness.

\subsection{Generalization}\label{sec4.4}
Fig.~\ref{teaser} demonstrates modality bias in LVLMs using LLaVA-1.5. To validate its generality, we analyzed Qwen-VL-Chat \cite{bai2023qwenvl}'s attention patterns about generative/discriminative hallucination tokens in Fig.~\ref{qwen_pattern}, revealing severe modality bias there as well. We also quantified attention patterns of hallucinated vs. non-hallucinated tokens (Fig. \ref{overall_pattern}), showing non-hallucinated ones have more balanced TAR/VAR with minimal bias. Given such pervasive modality bias, our TVAI can broadly enhance LVLMs' reliability.

\section{Conclusion}

In this study, we investigated the underlying mechanisms of hallucination in large vision-language models (LVLMs) and empirically identified a key contributing factor: \textit{modality bias}—a prevalent issue in current LVLMs that leads to over-reliance on either visual or textual modalities. To address this, we proposed TVAI, a simple yet effective training-free method for hallucination mitigation. TVAI guides LVLMs to better align their responses with user instructions, thereby enhancing response fidelity and reducing spurious content. Our experimental results highlight its potential as a practical tool for improving multimodal reasoning in LVLMs.

\bibliography{aaai2026}

\begin{thebibliography}{45}
\providecommand{\natexlab}[1]{#1}

\bibitem[{Achiam et~al.(2023)Achiam, Adler, Agarwal, Ahmad, Akkaya, Aleman, Almeida, Altenschmidt, Altman, Anadkat et~al.}]{achiam2023gpt}
Achiam, J.; Adler, S.; Agarwal, S.; Ahmad, L.; Akkaya, I.; Aleman, F.~L.; Almeida, D.; Altenschmidt, J.; Altman, S.; Anadkat, S.; et~al. 2023.
\newblock Gpt-4 technical report.
\newblock \emph{arXiv preprint arXiv:2303.08774}.

\bibitem[{An et~al.(2024)An, Tian, Leng, Nie, Lin, Wang, Dai, Chen, and Lu}]{an2024agla}
An, W.; Tian, F.; Leng, S.; Nie, J.; Lin, H.; Wang, Q.; Dai, G.; Chen, P.; and Lu, S. 2024.
\newblock Agla: Mitigating object hallucinations in large vision-language models with assembly of global and local attention.
\newblock \emph{arXiv preprint arXiv:2406.12718}.

\bibitem[{Bai et~al.(2023{\natexlab{a}})Bai, Bai, Yang, Wang, Tan, Wang, Lin, Zhou, and Zhou}]{bai2023qwen}
Bai, J.; Bai, S.; Yang, S.; Wang, S.; Tan, S.; Wang, P.; Lin, J.; Zhou, C.; and Zhou, J. 2023{\natexlab{a}}.
\newblock Qwen-vl: A versatile vision-language model for understanding, localization, text reading, and beyond.
\newblock \emph{arXiv preprint arXiv:2308.12966}, 1(2): 3.

\bibitem[{Bai et~al.(2023{\natexlab{b}})Bai, Bai, Yang, Wang, Tan, Wang, Lin, Zhou, and Zhou}]{bai2023qwenvl}
Bai, J.; Bai, S.; Yang, S.; Wang, S.; Tan, S.; Wang, P.; Lin, J.; Zhou, C.; and Zhou, J. 2023{\natexlab{b}}.
\newblock Qwen-vl: A versatile vision-language model for understanding, localization, text reading, and beyond.
\newblock \emph{arXiv preprint arXiv:2308.12966}, 1(2): 3.

\bibitem[{Brown et~al.(2020)Brown, Mann, Ryder, Subbiah, Kaplan, Dhariwal, Neelakantan, Shyam, Sastry, Askell et~al.}]{brown2020language}
Brown, T.; Mann, B.; Ryder, N.; Subbiah, M.; Kaplan, J.~D.; Dhariwal, P.; Neelakantan, A.; Shyam, P.; Sastry, G.; Askell, A.; et~al. 2020.
\newblock Language models are few-shot learners.
\newblock \emph{Advances in neural information processing systems}, 33: 1877--1901.

\bibitem[{Chen et~al.(2024{\natexlab{a}})Chen, Zhang, Huang, Niu, Zhang, Wen, and Hu}]{chen2024ict}
Chen, J.; Zhang, T.; Huang, S.; Niu, Y.; Zhang, L.; Wen, L.; and Hu, X. 2024{\natexlab{a}}.
\newblock ICT: Image-Object Cross-Level Trusted Intervention for Mitigating Object Hallucination in Large Vision-Language Models.
\newblock \emph{arXiv preprint arXiv:2411.15268}.

\bibitem[{Chen et~al.(2023)Chen, Zhang, Zeng, Zhang, Zhu, and Zhao}]{chen2023shikra}
Chen, K.; Zhang, Z.; Zeng, W.; Zhang, R.; Zhu, F.; and Zhao, R. 2023.
\newblock Shikra: Unleashing multimodal llm's referential dialogue magic.
\newblock \emph{arXiv preprint arXiv:2306.15195}.

\bibitem[{Chen et~al.(2024{\natexlab{b}})Chen, Zhao, Liu, Bai, Lin, Zhou, and Chang}]{chen2024image}
Chen, L.; Zhao, H.; Liu, T.; Bai, S.; Lin, J.; Zhou, C.; and Chang, B. 2024{\natexlab{b}}.
\newblock An image is worth 1/2 tokens after layer 2: Plug-and-play inference acceleration for large vision-language models.
\newblock In \emph{European Conference on Computer Vision}, 19--35. Springer.

\bibitem[{Chen et~al.(2024{\natexlab{c}})Chen, Zhao, Luo, Yao, Li, and Zhou}]{chen2024halc}
Chen, Z.; Zhao, Z.; Luo, H.; Yao, H.; Li, B.; and Zhou, J. 2024{\natexlab{c}}.
\newblock Halc: Object hallucination reduction via adaptive focal-contrast decoding.
\newblock \emph{arXiv preprint arXiv:2403.00425}.

\bibitem[{Chiang et~al.(2023)Chiang, Li, Lin, Sheng, Wu, Zhang, Zheng, Zhuang, Zhuang, Gonzalez et~al.}]{chiang2023vicuna}
Chiang, W.-L.; Li, Z.; Lin, Z.; Sheng, Y.; Wu, Z.; Zhang, H.; Zheng, L.; Zhuang, S.; Zhuang, Y.; Gonzalez, J.~E.; et~al. 2023.
\newblock Vicuna: An open-source chatbot impressing gpt-4 with 90\%* chatgpt quality.
\newblock \emph{See https://vicuna. lmsys. org (accessed 14 April 2023)}, 2(3): 6.

\bibitem[{Cho et~al.(2022)Cho, Yoon, Kale, Dernoncourt, Bui, and Bansal}]{cho2022fine}
Cho, J.; Yoon, S.; Kale, A.; Dernoncourt, F.; Bui, T.; and Bansal, M. 2022.
\newblock Fine-grained image captioning with clip reward.
\newblock \emph{arXiv preprint arXiv:2205.13115}.

\bibitem[{Chuang et~al.(2024)Chuang, Qiu, Hsieh, Krishna, Kim, and Glass}]{chuang2024lookback}
Chuang, Y.-S.; Qiu, L.; Hsieh, C.-Y.; Krishna, R.; Kim, Y.; and Glass, J. 2024.
\newblock Lookback lens: Detecting and mitigating contextual hallucinations in large language models using only attention maps.
\newblock \emph{arXiv preprint arXiv:2407.07071}.

\bibitem[{Chuang et~al.(2023)Chuang, Xie, Luo, Kim, Glass, and He}]{chuang2023dola}
Chuang, Y.-S.; Xie, Y.; Luo, H.; Kim, Y.; Glass, J.; and He, P. 2023.
\newblock Dola: Decoding by contrasting layers improves factuality in large language models.
\newblock \emph{arXiv preprint arXiv:2309.03883}.

\bibitem[{Cui et~al.(2024)Cui, Ma, Cao, Ye, Zhou, Liang, Chen, Lu, Yang, Liao et~al.}]{cui2024survey}
Cui, C.; Ma, Y.; Cao, X.; Ye, W.; Zhou, Y.; Liang, K.; Chen, J.; Lu, J.; Yang, Z.; Liao, K.-D.; et~al. 2024.
\newblock A survey on multimodal large language models for autonomous driving.
\newblock In \emph{Proceedings of the IEEE/CVF Winter Conference on Applications of Computer Vision}, 958--979.

\bibitem[{Darcet et~al.(2023)Darcet, Oquab, Mairal, and Bojanowski}]{darcet2023vision}
Darcet, T.; Oquab, M.; Mairal, J.; and Bojanowski, P. 2023.
\newblock Vision transformers need registers.
\newblock \emph{arXiv preprint arXiv:2309.16588}.

\bibitem[{Gunasekar et~al.(2023)Gunasekar, Zhang, Aneja, Mendes, Del~Giorno, Gopi, Javaheripi, Kauffmann, de~Rosa, Saarikivi et~al.}]{gunasekar2023textbooks}
Gunasekar, S.; Zhang, Y.; Aneja, J.; Mendes, C. C.~T.; Del~Giorno, A.; Gopi, S.; Javaheripi, M.; Kauffmann, P.; de~Rosa, G.; Saarikivi, O.; et~al. 2023.
\newblock Textbooks are all you need.
\newblock \emph{arXiv preprint arXiv:2306.11644}.

\bibitem[{Gunjal, Yin, and Bas(2024)}]{gunjal2024detecting}
Gunjal, A.; Yin, J.; and Bas, E. 2024.
\newblock Detecting and preventing hallucinations in large vision language models.
\newblock In \emph{Proceedings of the AAAI Conference on Artificial Intelligence}, volume~38, 18135--18143.

\bibitem[{Huang et~al.(2025)Huang, Yu, Ma, Zhong, Feng, Wang, Chen, Peng, Feng, Qin et~al.}]{huang2025survey}
Huang, L.; Yu, W.; Ma, W.; Zhong, W.; Feng, Z.; Wang, H.; Chen, Q.; Peng, W.; Feng, X.; Qin, B.; et~al. 2025.
\newblock A survey on hallucination in large language models: Principles, taxonomy, challenges, and open questions.
\newblock \emph{ACM Transactions on Information Systems}, 43(2): 1--55.

\bibitem[{Huang et~al.(2024)Huang, Dong, Zhang, Wang, He, Wang, Lin, Zhang, and Yu}]{huang2024opera}
Huang, Q.; Dong, X.; Zhang, P.; Wang, B.; He, C.; Wang, J.; Lin, D.; Zhang, W.; and Yu, N. 2024.
\newblock Opera: Alleviating hallucination in multi-modal large language models via over-trust penalty and retrospection-allocation.
\newblock In \emph{Proceedings of the IEEE/CVF Conference on Computer Vision and Pattern Recognition}, 13418--13427.

\bibitem[{Jiang et~al.(2024)Jiang, Chen, Zhu, Luo, Shen, and Yang}]{jiang2024devils}
Jiang, Z.; Chen, J.; Zhu, B.; Luo, T.; Shen, Y.; and Yang, X. 2024.
\newblock Devils in middle layers of large vision-language models: Interpreting, detecting and mitigating object hallucinations via attention lens.
\newblock \emph{arXiv preprint arXiv:2411.16724}.

\bibitem[{Lee et~al.(2021)Lee, Ippolito, Nystrom, Zhang, Eck, Callison-Burch, and Carlini}]{lee2021deduplicating}
Lee, K.; Ippolito, D.; Nystrom, A.; Zhang, C.; Eck, D.; Callison-Burch, C.; and Carlini, N. 2021.
\newblock Deduplicating training data makes language models better.
\newblock \emph{arXiv preprint arXiv:2107.06499}.

\bibitem[{Leng et~al.(2024)Leng, Zhang, Chen, Li, Lu, Miao, and Bing}]{leng2024mitigating}
Leng, S.; Zhang, H.; Chen, G.; Li, X.; Lu, S.; Miao, C.; and Bing, L. 2024.
\newblock Mitigating object hallucinations in large vision-language models through visual contrastive decoding.
\newblock In \emph{Proceedings of the IEEE/CVF Conference on Computer Vision and Pattern Recognition}, 13872--13882.

\bibitem[{Li et~al.(2024{\natexlab{a}})Li, Wong, Zhang, Usuyama, Liu, Yang, Naumann, Poon, and Gao}]{li2024llava}
Li, C.; Wong, C.; Zhang, S.; Usuyama, N.; Liu, H.; Yang, J.; Naumann, T.; Poon, H.; and Gao, J. 2024{\natexlab{a}}.
\newblock Llava-med: Training a large language-and-vision assistant for biomedicine in one day.
\newblock \emph{Advances in Neural Information Processing Systems}, 36.

\bibitem[{Li et~al.(2024{\natexlab{b}})Li, Xie, Li, Chen, Wang, Chen, Yang, Wang, Kong, and Liu}]{li2024vlfeedback}
Li, L.; Xie, Z.; Li, M.; Chen, S.; Wang, P.; Chen, L.; Yang, Y.; Wang, B.; Kong, L.; and Liu, Q. 2024{\natexlab{b}}.
\newblock VLFeedback: A Large-Scale AI Feedback Dataset for Large Vision-Language Models Alignment.
\newblock \emph{arXiv preprint arXiv:2410.09421}.

\bibitem[{Li et~al.(2023)Li, Du, Zhou, Wang, Zhao, and Wen}]{li2023evaluating}
Li, Y.; Du, Y.; Zhou, K.; Wang, J.; Zhao, W.~X.; and Wen, J.-R. 2023.
\newblock Evaluating object hallucination in large vision-language models.
\newblock \emph{arXiv preprint arXiv:2305.10355}.

\bibitem[{Lin et~al.(2014)Lin, Maire, Belongie, Hays, Perona, Ramanan, Doll{\'a}r, and Zitnick}]{lin2014microsoft}
Lin, T.-Y.; Maire, M.; Belongie, S.; Hays, J.; Perona, P.; Ramanan, D.; Doll{\'a}r, P.; and Zitnick, C.~L. 2014.
\newblock Microsoft coco: Common objects in context.
\newblock In \emph{Computer vision--ECCV 2014: 13th European conference, zurich, Switzerland, September 6-12, 2014, proceedings, part v 13}, 740--755. Springer.

\bibitem[{Liu et~al.(2024{\natexlab{a}})Liu, Li, Li, and Lee}]{liu2024improved}
Liu, H.; Li, C.; Li, Y.; and Lee, Y.~J. 2024{\natexlab{a}}.
\newblock Improved baselines with visual instruction tuning.
\newblock In \emph{Proceedings of the IEEE/CVF Conference on Computer Vision and Pattern Recognition}, 26296--26306.

\bibitem[{Liu et~al.(2023)Liu, Li, Wu, and Lee}]{liu2023visual}
Liu, H.; Li, C.; Wu, Q.; and Lee, Y.~J. 2023.
\newblock Visual instruction tuning.
\newblock \emph{Advances in neural information processing systems}, 36: 34892--34916.

\bibitem[{Liu et~al.(2024{\natexlab{b}})Liu, Xue, Chen, Chen, Zhao, Wang, Hou, Li, and Peng}]{liu2024survey}
Liu, H.; Xue, W.; Chen, Y.; Chen, D.; Zhao, X.; Wang, K.; Hou, L.; Li, R.; and Peng, W. 2024{\natexlab{b}}.
\newblock A survey on hallucination in large vision-language models.
\newblock \emph{arXiv preprint arXiv:2402.00253}.

\bibitem[{Liu et~al.(2024{\natexlab{c}})Liu, Ye, Xing, and Zou}]{liu2024reducing}
Liu, S.; Ye, H.; Xing, L.; and Zou, J. 2024{\natexlab{c}}.
\newblock Reducing hallucinations in vision-language models via latent space steering.
\newblock \emph{arXiv preprint arXiv:2410.15778}.

\bibitem[{Liu, Zheng, and Chen(2024)}]{liu2024paying}
Liu, S.; Zheng, K.; and Chen, W. 2024.
\newblock Paying more attention to image: A training-free method for alleviating hallucination in lvlms.
\newblock In \emph{European Conference on Computer Vision}, 125--140. Springer.

\bibitem[{Liu et~al.(2024{\natexlab{d}})Liu, Duan, Zhang, Li, Zhang, Zhao, Yuan, Wang, He, Liu et~al.}]{liu2024mmbench}
Liu, Y.; Duan, H.; Zhang, Y.; Li, B.; Zhang, S.; Zhao, W.; Yuan, Y.; Wang, J.; He, C.; Liu, Z.; et~al. 2024{\natexlab{d}}.
\newblock Mmbench: Is your multi-modal model an all-around player?
\newblock In \emph{European conference on computer vision}, 216--233. Springer.

\bibitem[{Peng et~al.(2023)Peng, Galley, He, Cheng, Xie, Hu, Huang, Liden, Yu, Chen et~al.}]{peng2023check}
Peng, B.; Galley, M.; He, P.; Cheng, H.; Xie, Y.; Hu, Y.; Huang, Q.; Liden, L.; Yu, Z.; Chen, W.; et~al. 2023.
\newblock Check your facts and try again: Improving large language models with external knowledge and automated feedback.
\newblock \emph{arXiv preprint arXiv:2302.12813}.

\bibitem[{Radford et~al.(2021)Radford, Kim, Hallacy, Ramesh, Goh, Agarwal, Sastry, Askell, Mishkin, Clark et~al.}]{radford2021learning}
Radford, A.; Kim, J.~W.; Hallacy, C.; Ramesh, A.; Goh, G.; Agarwal, S.; Sastry, G.; Askell, A.; Mishkin, P.; Clark, J.; et~al. 2021.
\newblock Learning transferable visual models from natural language supervision.
\newblock In \emph{International conference on machine learning}, 8748--8763. PmLR.

\bibitem[{Rohrbach et~al.(2018)Rohrbach, Hendricks, Burns, Darrell, and Saenko}]{rohrbach2018object}
Rohrbach, A.; Hendricks, L.~A.; Burns, K.; Darrell, T.; and Saenko, K. 2018.
\newblock Object hallucination in image captioning.
\newblock \emph{arXiv preprint arXiv:1809.02156}.

\bibitem[{Tian et~al.(2024)Tian, Gu, Li, Liu, Wang, Zhao, Zhan, Jia, Lang, and Zhao}]{tian2024drivevlm}
Tian, X.; Gu, J.; Li, B.; Liu, Y.; Wang, Y.; Zhao, Z.; Zhan, K.; Jia, P.; Lang, X.; and Zhao, H. 2024.
\newblock Drivevlm: The convergence of autonomous driving and large vision-language models.
\newblock \emph{arXiv preprint arXiv:2402.12289}.

\bibitem[{Touvron et~al.(2023)Touvron, Lavril, Izacard, Martinet, Lachaux, Lacroix, Rozi{\`e}re, Goyal, Hambro, Azhar et~al.}]{touvron2023llama}
Touvron, H.; Lavril, T.; Izacard, G.; Martinet, X.; Lachaux, M.-A.; Lacroix, T.; Rozi{\`e}re, B.; Goyal, N.; Hambro, E.; Azhar, F.; et~al. 2023.
\newblock Llama: Open and efficient foundation language models.
\newblock \emph{arXiv preprint arXiv:2302.13971}.

\bibitem[{Vaswani et~al.(2017)Vaswani, Shazeer, Parmar, Uszkoreit, Jones, Gomez, Kaiser, and Polosukhin}]{vaswani2017attention}
Vaswani, A.; Shazeer, N.; Parmar, N.; Uszkoreit, J.; Jones, L.; Gomez, A.~N.; Kaiser, {\L}.; and Polosukhin, I. 2017.
\newblock Attention is all you need.
\newblock \emph{Advances in neural information processing systems}, 30.

\bibitem[{Wang et~al.(2024)Wang, Chen, Zhang, Tian, Xu, Deng, and Chen}]{wang2024mllm}
Wang, C.; Chen, X.; Zhang, N.; Tian, B.; Xu, H.; Deng, S.; and Chen, H. 2024.
\newblock Mllm can see? dynamic correction decoding for hallucination mitigation.
\newblock \emph{arXiv preprint arXiv:2410.11779}.

\bibitem[{Wu et~al.(2024)Wu, Liu, Wang, Zhang, Wu, Wang, and Tan}]{wu2024logical}
Wu, J.; Liu, Q.; Wang, D.; Zhang, J.; Wu, S.; Wang, L.; and Tan, T. 2024.
\newblock Logical closed loop: Uncovering object hallucinations in large vision-language models.
\newblock \emph{arXiv preprint arXiv:2402.11622}.

\bibitem[{Xiao et~al.(2023)Xiao, Tian, Chen, Han, and Lewis}]{xiao2023efficient}
Xiao, G.; Tian, Y.; Chen, B.; Han, S.; and Lewis, M. 2023.
\newblock Efficient streaming language models with attention sinks.
\newblock \emph{arXiv preprint arXiv:2309.17453}.

\bibitem[{Xu et~al.(2024)Xu, Wang, Zhou, Ma, Yang, Lin, Wang, Wang, Liang, Han et~al.}]{xu2024multimodal}
Xu, Y.; Wang, Y.; Zhou, F.; Ma, J.; Yang, S.; Lin, H.; Wang, X.; Wang, J.; Liang, L.; Han, A.; et~al. 2024.
\newblock A multimodal knowledge-enhanced whole-slide pathology foundation model.
\newblock \emph{arXiv preprint arXiv:2407.15362}.

\bibitem[{Zhao et~al.(2024)Zhao, Si, Chen, Zhang, Sun, Zhang, and Chang}]{zhao2024looking}
Zhao, H.; Si, S.; Chen, L.; Zhang, Y.; Sun, M.; Zhang, M.; and Chang, B. 2024.
\newblock Looking Beyond Text: Reducing Language bias in Large Vision-Language Models via Multimodal Dual-Attention and Soft-Image Guidance.
\newblock \emph{arXiv preprint arXiv:2411.14279}.

\bibitem[{Zhou et~al.(2023)Zhou, Cui, Yoon, Zhang, Deng, Finn, Bansal, and Yao}]{zhou2023analyzing}
Zhou, Y.; Cui, C.; Yoon, J.; Zhang, L.; Deng, Z.; Finn, C.; Bansal, M.; and Yao, H. 2023.
\newblock Analyzing and mitigating object hallucination in large vision-language models.
\newblock \emph{arXiv preprint arXiv:2310.00754}.

\bibitem[{Zhu et~al.(2023)Zhu, Chen, Shen, Li, and Elhoseiny}]{zhu2023minigpt}
Zhu, D.; Chen, J.; Shen, X.; Li, X.; and Elhoseiny, M. 2023.
\newblock Minigpt-4: Enhancing vision-language understanding with advanced large language models.
\newblock \emph{arXiv preprint arXiv:2304.10592}.

\end{thebibliography}

\makeatletter
\@ifundefined{isChecklistMainFile}{
  \newif\ifreproStandalone
  \reproStandalonetrue
}{
  \newif\ifreproStandalone
  \reproStandalonefalse
}
\makeatother

\ifreproStandalone
\documentclass[letterpaper]{article}
\usepackage[submission]{aaai2026}
\setlength{\pdfpagewidth}{8.5in}
\setlength{\pdfpageheight}{11in}
\usepackage{times}
\usepackage{helvet}
\usepackage{courier}
\usepackage{xcolor}
\frenchspacing

\begin{document}
\fi
\setlength{\leftmargini}{20pt}
\makeatletter\def\@listi{\leftmargin\leftmargini \topsep .5em \parsep .5em \itemsep .5em}
\def\@listii{\leftmargin\leftmarginii \labelwidth\leftmarginii \advance\labelwidth-\labelsep \topsep .4em \parsep .4em \itemsep .4em}
\def\@listiii{\leftmargin\leftmarginiii \labelwidth\leftmarginiii \advance\labelwidth-\labelsep \topsep .4em \parsep .4em \itemsep .4em}\makeatother

\setcounter{secnumdepth}{0}
\renewcommand\thesubsection{\arabic{subsection}}
\renewcommand\labelenumi{\thesubsection.\arabic{enumi}}

\newcounter{checksubsection}
\newcounter{checkitem}[checksubsection]

\newcommand{\checksubsection}[1]{%
  \refstepcounter{checksubsection}%
  \paragraph{\arabic{checksubsection}. #1}%
  \setcounter{checkitem}{0}%
}

\newcommand{\checkitem}{%
  \refstepcounter{checkitem}%
  \item[\arabic{checksubsection}.\arabic{checkitem}.]%
}
\newcommand{\question}[2]{\normalcolor\checkitem #1 #2 \color{blue}}
\newcommand{\ifyespoints}[1]{\makebox[0pt][l]{\hspace{-15pt}\normalcolor #1}}

\section*{Reproducibility Checklist}

\vspace{1em}
\hrule
\vspace{1em}

\textbf{Instructions for Authors:}

This document outlines key aspects for assessing reproducibility. Please provide your input by editing this \texttt{.tex} file directly.

For each question (that applies), replace the ``Type your response here'' text with your answer.

\vspace{1em}
\noindent
\textbf{Example:} If a question appears as
\begin{center}
\noindent
\begin{minipage}{.9\linewidth}
\ttfamily\raggedright
\string\question \{Proofs of all novel claims are included\} \{(yes/partial/no)\} \\
Type your response here
\end{minipage}
\end{center}
you would change it to:
\begin{center}
\noindent
\begin{minipage}{.9\linewidth}
\ttfamily\raggedright
\string\question \{Proofs of all novel claims are included\} \{(yes/partial/no)\} \\
yes
\end{minipage}
\end{center}
Please make sure to:
\begin{itemize}\setlength{\itemsep}{.1em}
\item Replace ONLY the ``Type your response here'' text and nothing else.
\item Use one of the options listed for that question (e.g., \textbf{yes}, \textbf{no}, \textbf{partial}, or \textbf{NA}).
\item \textbf{Not} modify any other part of the \texttt{\string\question} command or any other lines in this document.\\
\end{itemize}

You can \texttt{\string\input} this .tex file right before \texttt{\string\end\{document\}} of your main file or compile it as a stand-alone document. Check the instructions on your conference's website to see if you will be asked to provide this checklist with your paper or separately.

\vspace{1em}
\hrule
\vspace{1em}


\checksubsection{General Paper Structure}
\begin{itemize}

\question{Includes a conceptual outline and/or pseudocode description of AI methods introduced}{(yes/partial/no/NA)}
yes

\question{Clearly delineates statements that are opinions, hypothesis, and speculation from objective facts and results}{(yes/no)}
yes

\question{Provides well-marked pedagogical references for less-familiar readers to gain background necessary to replicate the paper}{(yes/no)}
yes

\end{itemize}
\checksubsection{Theoretical Contributions}
\begin{itemize}

\question{Does this paper make theoretical contributions?}{(yes/no)}
no

	\ifyespoints{\vspace{1.2em}If yes, please address the following points:}
        \begin{itemize}
	
	\question{All assumptions and restrictions are stated clearly and formally}{(yes/partial/no)}
	NA

	\question{All novel claims are stated formally (e.g., in theorem statements)}{(yes/partial/no)}
	NA

	\question{Proofs of all novel claims are included}{(yes/partial/no)}
	NA

	\question{Proof sketches or intuitions are given for complex and/or novel results}{(yes/partial/no)}
	NA

	\question{Appropriate citations to theoretical tools used are given}{(yes/partial/no)}
	NA

	\question{All theoretical claims are demonstrated empirically to hold}{(yes/partial/no/NA)}
	NA

	\question{All experimental code used to eliminate or disprove claims is included}{(yes/no/NA)}
	NA
	
	\end{itemize}
\end{itemize}

\checksubsection{Dataset Usage}
\begin{itemize}

\question{Does this paper rely on one or more datasets?}{(yes/no)}
yes

\ifyespoints{If yes, please address the following points:}
\begin{itemize}

	\question{A motivation is given for why the experiments are conducted on the selected datasets}{(yes/partial/no/NA)}
	yes

	\question{All novel datasets introduced in this paper are included in a data appendix}{(yes/partial/no/NA)}
	yes

	\question{All novel datasets introduced in this paper will be made publicly available upon publication of the paper with a license that allows free usage for research purposes}{(yes/partial/no/NA)}
	yes

	\question{All datasets drawn from the existing literature (potentially including authors' own previously published work) are accompanied by appropriate citations}{(yes/no/NA)}
	yes

	\question{All datasets drawn from the existing literature (potentially including authors' own previously published work) are publicly available}{(yes/partial/no/NA)}
	yes

	\question{All datasets that are not publicly available are described in detail, with explanation why publicly available alternatives are not scientifically satisficing}{(yes/partial/no/NA)}
        NA

\end{itemize}
\end{itemize}

\checksubsection{Computational Experiments}
\begin{itemize}

\question{Does this paper include computational experiments?}{(yes/no)}
yes

\ifyespoints{If yes, please address the following points:}
\begin{itemize}

	\question{This paper states the number and range of values tried per (hyper-) parameter during development of the paper, along with the criterion used for selecting the final parameter setting}{(yes/partial/no/NA)}
	yes

	\question{Any code required for pre-processing data is included in the appendix}{(yes/partial/no)}
	yes

	\question{All source code required for conducting and analyzing the experiments is included in a code appendix}{(yes/partial/no)}
	yes

	\question{All source code required for conducting and analyzing the experiments will be made publicly available upon publication of the paper with a license that allows free usage for research purposes}{(yes/partial/no)}
	yes
        
	\question{All source code implementing new methods have comments detailing the implementation, with references to the paper where each step comes from}{(yes/partial/no)}
	yes

	\question{If an algorithm depends on randomness, then the method used for setting seeds is described in a way sufficient to allow replication of results}{(yes/partial/no/NA)}
	yes

	\question{This paper specifies the computing infrastructure used for running experiments (hardware and software), including GPU/CPU models; amount of memory; operating system; names and versions of relevant software libraries and frameworks}{(yes/partial/no)}
	yes

	\question{This paper formally describes evaluation metrics used and explains the motivation for choosing these metrics}{(yes/partial/no)}
	yes

	\question{This paper states the number of algorithm runs used to compute each reported result}{(yes/no)}
	yes

	\question{Analysis of experiments goes beyond single-dimensional summaries of performance (e.g., average; median) to include measures of variation, confidence, or other distributional information}{(yes/no)}
	yes

	\question{The significance of any improvement or decrease in performance is judged using appropriate statistical tests (e.g., Wilcoxon signed-rank)}{(yes/partial/no)}
	yes

	\question{This paper lists all final (hyper-)parameters used for each model/algorithm in the paper’s experiments}{(yes/partial/no/NA)}
	yes

\end{itemize}
\end{itemize}
\ifreproStandalone
\end{document}
\fi

\clearpage
\appendix

\onecolumn
{\centering
\Large
\textbf{Modality Bias in LVLMs: \\ Analyzing and Mitigating Object Hallucination via Attention Lens}\\
}
\section{Supplementary Material}

\subsection{Related Work}
Large language models (LLMs) \cite{achiam2023gpt,brown2020language} have experienced rapid advancements in recent years, catalyzing extensive research into multimodal large language models (MLLMs) within the AI community. Supported by open-source LLMs such as LLaMA \cite{touvron2023llama}, Qwen \cite{bai2023qwen}, and Vicuna \cite{chiang2023vicuna}, a large number of LVLMs have been developed and proposed \cite{liu2023visual,bai2023qwenvl,zhu2023minigpt,chen2023shikra}. LVLMs transform image inputs into textual representation spaces within LLMs through visual encoders and projectors, seamlessly integrating multimodal information. Leveraging the advanced reasoning capabilities of LLMs, LVLMs effectively fulfill users' requirements. LLaVA-1.5 \cite{liu2024improved}, Qwen-VL \cite{bai2023qwenvl}, MiniGPT-4 \cite{zhu2023minigpt}, and Shikra \cite{chen2023shikra} are among the most prominent and widely adopted open-source LVLMs. These models generally adhere to a consistent training paradigm, which involves pre-training for modality alignment followed by instruction fine-tuning, enabling them to achieve robust multimodal instruction understanding and execution capabilities. Shikra directly processes both visual features and textual coordinate information through natural language sequences, thereby eliminating the need for staged alignment of multimodal representations. However, all the aforementioned LVLMs continue to exhibit severe hallucinated issues. As a result, we primarily conduct our experiments using these four widely adopted open-source models.

\subsection{Implementation Details}
In our experiments, we use the 7B parameter versions of all LVLMs under evaluation. For decoding, we adopt greedy search as the default strategy. However, since OPERA \cite{huang2024opera} is specifically designed to operate under beam search, we set the beam size to 5, following the configuration in \cite{liu2024paying}. Following PAI \cite{liu2024paying}, we conduct CHAIR evaluations on 500 randomly sampled images from the COCO 2014 validation set. To ensure reproducibility and clarity of our experimental framework, key configuration details are summarized in Table~\ref{tab:experimental_setup}. Specifically, all experiments are conducted with a fixed random seed of 2025 to stabilize stochastic processes. The maximum length of generated sequences is constrained to 512 tokens to balance output completeness and computational efficiency. Hardware-wise, experiments run on a single NVIDIA GeForce RTX 4090 GPU, providing sufficient computational power for model inference. To mitigate the "repetition problem" (i.e., redundant text generation), a post-processing strategy with an n-gram parameter of 4 is adopted, effectively suppressing consecutive repeated sequences. These configurations collectively ensure the reliability and comparability of our experimental results.

\begin{table}[h]
\centering

\begin{tabular}{lc}
\toprule
Configuration Item & Value \\
\midrule
Random Seed & 2025 \\
Maximum Generated Tokens & 512 \\
N-gram Parameter for Repetition Penalty & 4 \\
\bottomrule
\end{tabular}
\caption{Experimental configuration details}
\label{tab:experimental_setup}
\end{table}

\noindent During the computation of Text Attention Ratio (TAR) and Visual Attention Ratio (VAR) for individual words, a critical consideration arises: due to subword tokenization mechanisms commonly used in large language models, a single natural language word (e.g., compound terms or polysyllabic words) is often split into multiple subword tokens during preprocessing. For example, a complex word may be tokenized into 2–3 subword units to fit the model’s vocabulary. In such cases, directly associating the word with a single token introduces inaccuracies in attention quantification. To address this, we adopt a token-averaging strategy: for each target word, we first extract the TAR and VAR values of all subword tokens that compose it (where TAR measures the total attention weight allocated to all input text tokens, and VAR measures the total attention weight allocated to all input visual tokens). We then compute the arithmetic mean of these token-level TAR and VAR values to derive the final representative TAR and VAR for the entire word. This approach ensures that the attention metrics for a word reflect the collective contribution of all its constituent tokens, thereby enhancing the robustness and accuracy of our modality bias analysis.

\noindent Our TVAI architecture comprises three hyperparameters: $\alpha$, $\beta$, and $\gamma$, which respectively denote the manipulation strength of attention allocated to visual tokens and textual tokens in user instructions, as well as the strength of contrast decoding to enhance attention manipulation. Given that datasets such as CHAIR \cite{rohrbach2018object}, POPE \cite{li2023evaluating}, and MMBench \cite{liu2024mmbench} have small sample sizes and highly imbalanced distributions, different hyperparameter configurations are required to adapt to these small datasets, with specific settings detailed in Table~\ref{chair_hyperparameter}, \ref{pope_hyperparameter}, and \ref{mmbench_hyperparameter}. However, for large-scale datasets with generally consistent data distributions, a unified set of TVAI hyperparameters can be applied to achieve effective hallucination mitigation while preserving the original general capabilities of the model. Additionally, optimal TVAI hyperparameters vary across different model architectures, training paradigms, and training data; thus, adjustments should be made in specific applications. Notably, many other hallucination mitigation methods also require re-finetuning for different domains, data distributions, and model architectures. In this context, our training-free TVAI method demonstrates its advantage: it achieves hallucination mitigation performance comparable to supervised finetuning without the need for expensive data annotation or training.

\begin{table}[h]
\centering
\begin{tabular}{c|ccc}
    \toprule
     Method & $\alpha$ & $\beta$ & $\gamma$ \\
    \midrule
    LLaVA-1.5 & 0.93 & 0.5 & 1.1 \\
    MiniGPT-4 & 0.8 & 0.4 & 1.1 \\
    Shikra & 0.8 & 0.6 & 1.2 \\
    \bottomrule
\end{tabular}
\caption{Hyperparameter configurations in CHAIR for LLaVA-1.5, MiniGPT-4, and Shikra.}
\label{chair_hyperparameter}
\end{table}

\begin{table}[h]
\centering
\begin{tabular}{c|ccc}
    \toprule
     Method & $\alpha$ & $\beta$ & $\gamma$ \\
    \midrule
    LLaVA-1.5 & 0.93 & 0.5 & 1.2 \\
    MiniGPT-4 & 0.1 & 0.1 & 1.1 \\
    Shikra & 0.8 & 0.6 & 1.1 \\
    \bottomrule
\end{tabular}
\caption{Hyperparameter configurations in POPE for LLaVA-1.5, MiniGPT-4, and Shikra.}
\label{pope_hyperparameter}
\end{table}

\begin{table}[h]
\centering
\begin{tabular}{c|ccc}
    \toprule
     Method & $\alpha$ & $\beta$ & $\gamma$ \\
    \midrule
    LLaVA-1.5 & 0.5 & 0.5 & 1.1 \\
    \bottomrule
\end{tabular}
\caption{Hyperparameter configurations in MMBench for LLaVA-1.5.}
\label{mmbench_hyperparameter}
\end{table}

\subsection{Prompt Template}
To ensure consistent and task-aligned interactions with the model across different evaluation scenarios, we designed specific prompts tailored to the characteristics of each dataset. The prompts are structured to guide the model toward generating outputs compatible with the dataset’s task requirements (e.g., descriptive answers, binary judgments, or option selection). Table~\ref{tab:prompts} summarizes the key prompts used for each dataset.  

\noindent \textbf{CHAIR Dataset}. Focuses on image captioning tasks, so the prompt is designed to elicit detailed visual descriptions from the model, ensuring comprehensive coverage of image content.  

\noindent \textbf{POPE Dataset}. Targets object presence verification, where the prompt explicitly queries the existence of a specific object (denoted by \textless object\textgreater, replaced by the actual object name in practice).  

\noindent \textbf{MMBench Dataset}. Includes both English (EN) and Chinese (CN) sub-datasets for multi-choice question answering. The prompts emphasize selecting the best answer from predefined options (A–D) and formatting the output as a single letter, aligning with the dataset’s evaluation criteria. Note that only the English prompt is displayed in Table~\ref{tab:prompts} due to restrictions on Chinese characters in the current template. The full Chinese prompt, as used in our experiments, is available in our open-source code repository\footnote{\url{https://github.com/zhenghh8/TVAI}} for reference.  

\noindent These prompts are critical for standardizing the model’s input conditions and ensuring that generated outputs can be directly evaluated against the dataset’s ground-truth annotations.

\begin{table}[h]
\centering
\begin{tabular}{l p{13cm}}
\toprule
Dataset & Prompt Template \\
\midrule
CHAIR & ``Please help me describe the image in detail.'' \\
POPE & ``Is there a \textless object\textgreater  in the image?'' \\
MMBench (EN) & ``\textless hint\textgreater  \textless question\textgreater  Choose the best answer from the options \{A, B, C, D\} and indicate it with a letter (e.g., A).'' \\
\bottomrule
\end{tabular}
\caption{Prompt templates used for different datasets.}
\label{tab:prompts}
\end{table}

\subsection{Limitation}
Our TVAI method employs static intervention on attention distributions during inference, which is limited by different data distributions—similar to the challenges faced by supervised fine-tuning. In fact, a dynamic attention distribution manipulation method would better adapt to diverse model architectures and domains. By assessing the attention distribution of the currently generated token, it could dynamically adjust the direction and strength of manipulation in real-time, which constitutes our future research direction. Furthermore, our method is based on the observation that current LVLMs exhibit \textit{modality bias}, and we verify in the main text that this issue is widespread across different LVLMs. This bias may stem from disparities in architectural parameters for processing different modalities in current multimodal large models, imbalances in data volume across modalities, and the dominance of text generation during training. TVAI addresses modality bias by manipulating attention distributions across modalities within the transformer architecture, but it can only partially mitigate the issues caused by data preparation, model architecture, and training methods in current LVLMs. In the future, improved multimodal data preparation pipelines and training paradigms will be needed to fundamentally resolve this problem. We hope that our observations on \textit{modality bias} will inspire further research on multimodal large models.

\subsection{Visualization}

\begin{figure}[h]
    \centering
    \includegraphics[width=0.95\textwidth]{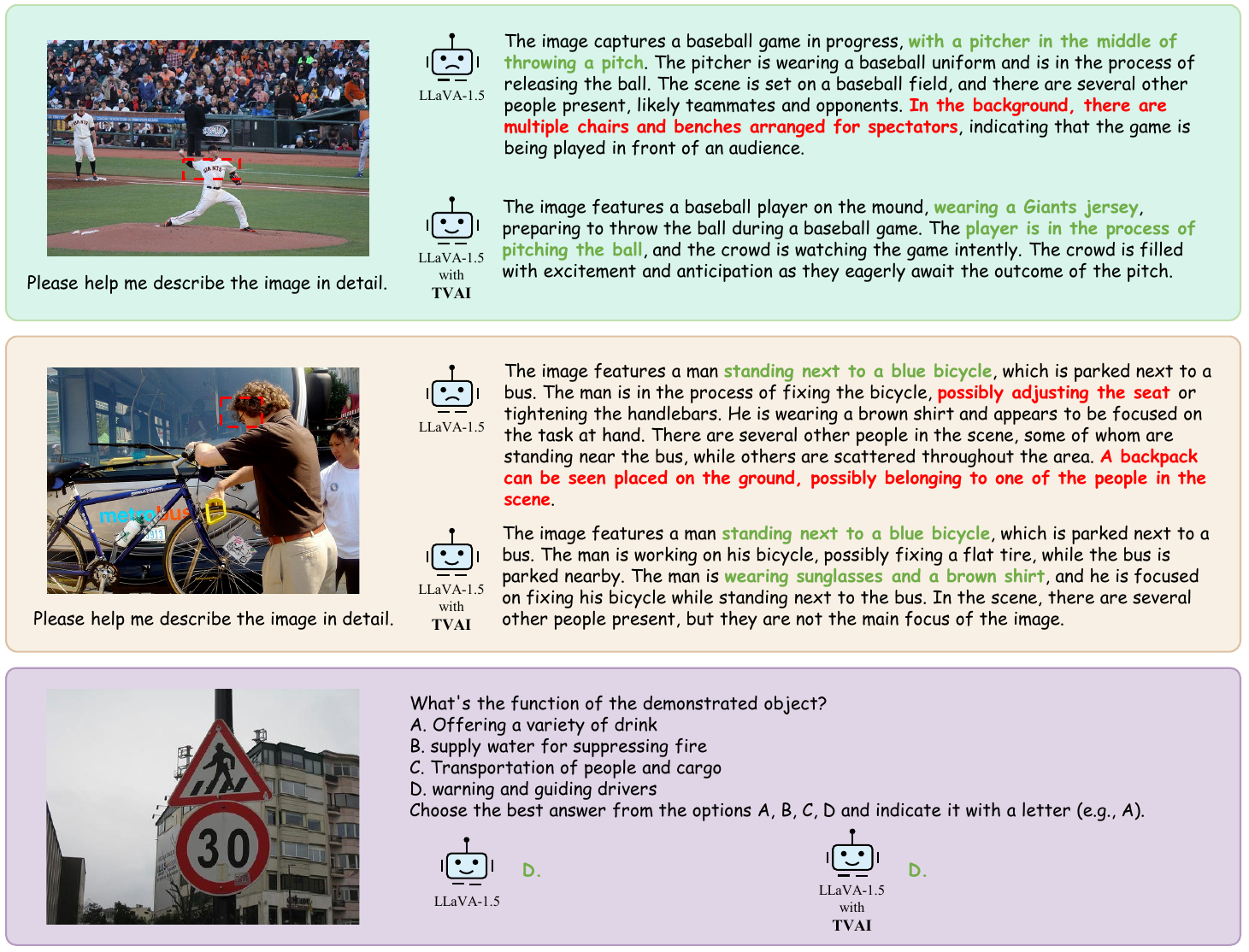}
    \caption{Some cases of LLaVA-1.5 with and without TVAI. Green bold text indicates correct descriptions, red bold text highlights hallucinations, and red dashed boxes in images mark fine-grained perceptual enhancements by TVAI.}
    \label{fig:visualization}
\end{figure}

Fig.~\ref{fig:visualization} illustrates three comparative examples of LLaVA-1.5's outputs with and without the integration of TVAI, highlighting the critical improvements introduced by our method.  

\noindent First, TVAI effectively mitigates hallucination, a common issue in multi-modal models where generated content deviates from the actual image or user input. In some cases, outputs without TVAI contain spurious details unsupported by the image (e.g., fabricating non-existent objects or actions), while TVAI-equipped models strictly align with visual content, ensuring factual consistency. This confirms TVAI's core capability to suppress hallucinatory generation by grounding the model's reasoning in real image cues. 

\noindent Beyond hallucination mitigation, our visualization reveals an additional key advantage: TVAI enhances the model's fine-grained perception ability, enabling it to capture subtle visual details that are overlooked in the baseline. For instance, in Example 1, the TVAI-augmented response precisely identifies that ``\textit{the baseball player is wearing a Giants jersey}''—a specific detail absent in the baseline, which only provides a generic description of the player. In Example 2, while the baseline merely notes the man's attire in vague terms, the TVAI-enhanced output specifies ``\textit{The man is wearing sunglasses and a brown shirt}'' accurately capturing both accessory and clothing details. These improvements demonstrate that TVAI empowers the model to discern and articulate nuanced visual features that are critical for comprehensive image understanding.  

\noindent We attribute these gains to TVAI's mechanism of rebalancing modal attention: by prompting the model to equally prioritize and attend to all modal information in user instructions (e.g., aligning textual queries with visual content more rigorously), TVAI prevents over-reliance on dominant modalities or prior biases. This balanced attention ensures that both prominent and subtle visual cues are properly processed, thereby reducing hallucinations and enhancing the model's capacity for fine-grained description. Collectively, these results validate that TVAI not only addresses a core limitation of multi-modal models but also unlocks enhanced perceptual capabilities by refining modal attention dynamics.

\end{document}